%% file: main.tex
\documentclass[letterpaper, 10 pt, conference]{IEEEconf}
\usepackage[letterpaper, left=0.75in, right=0.75in, bottom=0.8in, top=0.75in]{geometry} 
\IEEEoverridecommandlockouts                              
\def\BibTeX{{\rm B\kern-.05em{\sc i\kern-.025em b}\kern-.08em
    T\kern-.1667em\lower.7ex\hbox{E}\kern-.125emX}}
\usepackage[utf8]{inputenc}
\usepackage{balance}

\def\BibTeX{{\rm B\kern-.05em{\sc i\kern-.025em b}\kern-.08em
    T\kern-.1667em\lower.7ex\hbox{E}\kern-.125emX}}
\usepackage{cite}
\usepackage{amsthm}
\usepackage{multirow}
\usepackage[table,xcdraw]{xcolor}
\usepackage{algorithm,algpseudocode}
\usepackage{hyperref}
\usepackage{url}
\usepackage{xspace}
\usepackage{graphicx}
\usepackage{amsfonts}
\usepackage{amsmath,amssymb,amstext}
\usepackage{textcomp}
\usepackage{acro}
\usepackage{pgfplots}
\usepackage{pgfplotstable}
\pgfplotsset{compat=newest}
\usepgfplotslibrary{groupplots,dateplot}
\usepackage{soul}
\usepackage{mathtools,lipsum, nccmath,cuted,amsmath}
\usepackage[capitalise]{cleveref}
\usepackage{siunitx}
\usepackage{textcomp}

\usepackage{comment}
\usepackage{makecell}
\usepackage{algorithm}
\usepackage{gensymb}
\usepackage[normalem]{ulem}
\usepackage{cancel}
\usepackage{subcaption}

\newtheorem*{problem}{Problem}
\newtheorem*{example}{Example}
\let\oldexample\example
\renewcommand{\example}{\oldexample\normalfont}
\usepackage{listings}%
\definecolor{keywords}{HTML}{8A4A0B}

\definecolor{background}{HTML}{EEEEEE}

\definecolor{comments}{HTML}{868686}

\lstdefinelanguage{sdl}{
 morekeywords={scene,long,type,entity,'{','}', class},
 keywordstyle=\color{keywords},
    basicstyle=\scriptsize\ttfamily,
 morecomment=[l]{//}, 
 morecomment=[s]{/*}{*/}, 
 morestring=[b]",
    basicstyle=\scriptsize\ttfamily,%
 commentstyle=\color{comments}\ttfamily,
numbers=right,
    numberstyle=\scriptsize,
    stepnumber=1,
    numbersep=8pt,
breaklines=true,
    frame=tb,
 tabsize=4}
 
\lstdefinelanguage{specification}{
 morekeywords={Scenario,town,track, regions,weather,Constraints,Infraction_Metrics, Record, Frequency,'{','}', class, traffic_density,pedestrian_density},
 keywordstyle=\color{keywords},
    basicstyle=\scriptsize\ttfamily,
 morecomment=[l]{//}, 
 morecomment=[s]{/*}{*/}, 
 morestring=[b]",
    basicstyle=\scriptsize\ttfamily,%
 commentstyle=\color{comments}\ttfamily,
numbers=right,
    numberstyle=\scriptsize,
    stepnumber=1,
    numbersep=8pt,
breaklines=true,
    frame=tb,
 tabsize=4}

\usetikzlibrary{pgfplots.statistics, pgfplots.colorbrewer,pgfplots.dateplot,
        shadows,
        matrix} 

\definecolor{blueLine}{RGB}{57,106,177}
\definecolor{blueFill}{RGB}{114,147,203}
\definecolor{redLine}{RGB}{204,37,41}
\definecolor{greenLine}{RGB}{0,250,0}
\definecolor{blackLine}{RGB}{0,0,0}
\definecolor{goldLine}{RGB}{160,82,45}
\definecolor{brightgreen}{rgb}{0.4, 1.0, 0.0}
\definecolor{brinkpink}{rgb}{0.98, 0.38, 0.5}
\definecolor{cadmiumyellow}{rgb}{1.0, 0.96, 0.0}
\definecolor{cinnamon}{rgb}{0.82, 0.41, 0.12}
\definecolor{darkorange}{rgb}{1.0, 0.55, 0.0}
\definecolor{darkspringgreen}{rgb}{0.09, 0.45, 0.27}

\usepackage[font=footnotesize]{caption}
\usepackage{soul}
\usepackage{xcolor}
\usepackage{multirow}
\usepackage{soul}

\author{Shreyas Ramakrishna$^{1}$\textsuperscript{\textsection}, Baiting Luo$^{1}$\textsuperscript{\textsection}, Christopher B. Kuhn$^{2}$, Gabor Karsai$^{1}$, and Abhishek Dubey$^{1}$
\thanks{$^{1}$Institute for Software Integrated Systems, Vanderbilt University,  Nashville, TN 37212, USA}%
\thanks{$^{2}$Department of Electrical and Computer Engineering, Technical University of Munich, 80333 Munich, Germany}%
}

\DeclareAcronym{tfpg}{
  short = TFPG,
  long  = Timed Failure Propagation Graph,
}
\DeclareAcronym{ac}{
  short = AC,
  long  = Assurance Case,
}

\DeclareAcronym{bo}{
  short = BO,
  long  = Bayesian Optimization,
}

\DeclareAcronym{dnn}{
  short = DNN,
  long  = Deep Neural Network,
}

\DeclareAcronym{gbo}{
  short = GBO,
  long  = Guided Bayesian Optimization,
}

\DeclareAcronym{rns}{
  short = RNS,
  long  = Random Neighborhood Search,
}

\DeclareAcronym{cps}{
  short = CPS,
  long = Cyber Physical System,
}

\DeclareAcronym{ml}{
  short = ML,
  long = Machine Learning,
}

\DeclareAcronym{lbc}{
  short = LBC,
  long = Learning By Cheating,
}

\DeclareAcronym{adas}{
  short = ADAS,
  long = Advanced Driver Assistance Systems,
}

\DeclareAcronym{ai}{
  short = AI,
  long = Artificial Intelligence,
}

\DeclareAcronym{aebs}{
  short = AEBS,
  long = Automatic Emergency Braking System,
}
\DeclareAcronym{sdl}{
  short = SDL,
  long = Scenario Description Language
}

\DeclareAcronym{av}{
  short = AV,
  long = Autonomous Vehicle
}

\DeclareAcronym{resonate}{
  short = ReSonAte,
  long  = Runtime Safety Evaluation in Autonomous Systems,
}

\DeclareAcronym{dsml}{
  short = DSML,
  long  = Domain-Specific Modeling Language,
}
\DeclareAcronym{lec}{
  short = LEC,
  long = Learning Enabled Component,
}

\DeclareAcronym{cnn}{
  short = CNN,
  long = Convolutional Neural Network,
}

\DeclareAcronym{btd}{
  short = BTD,
  long  = Bow-Tie Diagram,
}

\DeclareAcronym{ood}{
  short = OOD,
  long = Out-of-Distribution
}

\DeclareAcronym{vae}{
  short = VAE,
  long = Variational Autoencoder,
}

\DeclareAcronym{gan}{
  short = GAN,
  long = Generative Adversarial Network,
}

\DeclareAcronym{gps}{
  short = GPS,
  long = Global Positioning System
}
\DeclareAcronym{imu}{
  short = IMU,
  long = Inertial Measurement Unit
}

\DeclareAcronym{rl}{
  short = RL,
  long = Reinforcement Learning
}

\DeclareAcronym{mdp}{
  short = MDP,
  long = Markov Decision Process
}

\DeclareAcronym{ucb}{
  short = UCB,
  long = Upper Confidence Bound
}
\DeclareAcronym{lut}{
  short = LUT,
  long = Lookup Table
}

\DeclareAcronym{gp}{
  short = GP,
  long = Gaussian Process
}

\usepackage{todonotes}

\begin{document}
\bstctlcite{IEEEexample:BSTcontrol}
  
\title{\LARGE \bf ANTI-CARLA: An Adversarial Testing Framework for Autonomous Vehicles in CARLA}

\maketitle



\begingroup\renewcommand\thefootnote{\textsection}
\footnotetext{These authors contributed equally to this work.}
\endgroup

\setcounter{page}{1}
\input{abstract.tex}
\pagestyle{plain}
\input{individual-sections/introduction}

\input{individual-sections/related-works}

\input{individual-sections/problem-statement}

\input{individual-sections/proposed-framework}

\input{individual-sections/evaluation}
\input{individual-sections/conclusion}

\balance
\bibliographystyle{IEEEtran}
\bibliography{main.bib}

\end{document}

%% file: abstract.tex
\begin{abstract}
Despite recent advances in autonomous driving systems, accidents such as the fatal Uber crash in 2018 show these systems are still susceptible to edge cases. Such systems must be thoroughly tested and validated before being deployed in the real world to avoid such events. Testing in open-world scenarios can be difficult, time-consuming, and expensive. These challenges can be addressed by using driving simulators such as CARLA instead. A key part of such tests is adversarial testing, in which the goal is to find scenarios that lead to failures of the given system. While several independent efforts in testing have been made, a well-established testing framework that enables adversarial testing has yet to be made available for CARLA. We therefore propose ANTI-CARLA, an automated testing framework in CARLA for simulating adversarial weather conditions (e.g., heavy rain) and sensor faults (e.g., camera occlusion) that fail the system. The operating conditions in which a given system should be tested are specified in a scenario description language. The framework offers an efficient search mechanism that searches for adversarial operating conditions that will fail the tested system. In this way, ANTI-CARLA extends the CARLA simulator with the capability of performing adversarial testing on any given driving pipeline. We use ANTI-CARLA to test the driving pipeline trained with Learning By Cheating (LBC) approach. The simulation results demonstrate that ANTI-CARLA can effectively and automatically find a range of failure cases despite LBC reaching an accuracy of 100\% in the CARLA benchmark. 
\end{abstract}


%% file: individual-sections/introduction.tex
\section{Introduction} 
\label{sec:intro} 
Advancements in \ac{ml} have led to considerable progress in the levels of autonomy achieved by \acp{av}~\cite{gibbs2017google}. 
However, recent accidents such as Tesla's autopilot crashes~\cite{vlasic2016self} and the fatal Uber self-driving car accident~\cite{kohli2019enabling} demonstrate that current \ac{av} systems still can fail. To address this, testing and validating such systems before deploying them into real-world operations has received increasing attention. To determine failure cases of a system before they happen on the road, it is crucial to generate adversarial test cases that can fail the system. For example, in our previous work~\cite{hartsell2021resonate,ramakrishna2022riskaware,habermayr2021situation}, we observed that an operational scene with adverse weather conditions (e.g., complex lighting, heavy precipitation, heavy fog) could be an adversarial test case for a vehicle operating based on visual perception. However, testing the system under such conditions in the real world can be expensive, slow, and often infeasible. Hence, recent research has been increasingly focusing on using synthetic data from simulators such as CARLA~\cite{dosovitskiy2017carla}, AirSim~\cite{shah2018airsim}, LGSVL~\cite{rong2020lgsvl}, and Deepdrive~\cite{deepdrive}. CARLA is particularly focused on autonomous driving and is a widely used option in academia~\cite{hofbauer2020telecarla,hartsell2021resonate,kuhn2021pixel,ramakrishna2022riskaware,hofbauer2020multi}. 

\begin{figure}[t] 
\centering 
\includegraphics[width=\columnwidth]{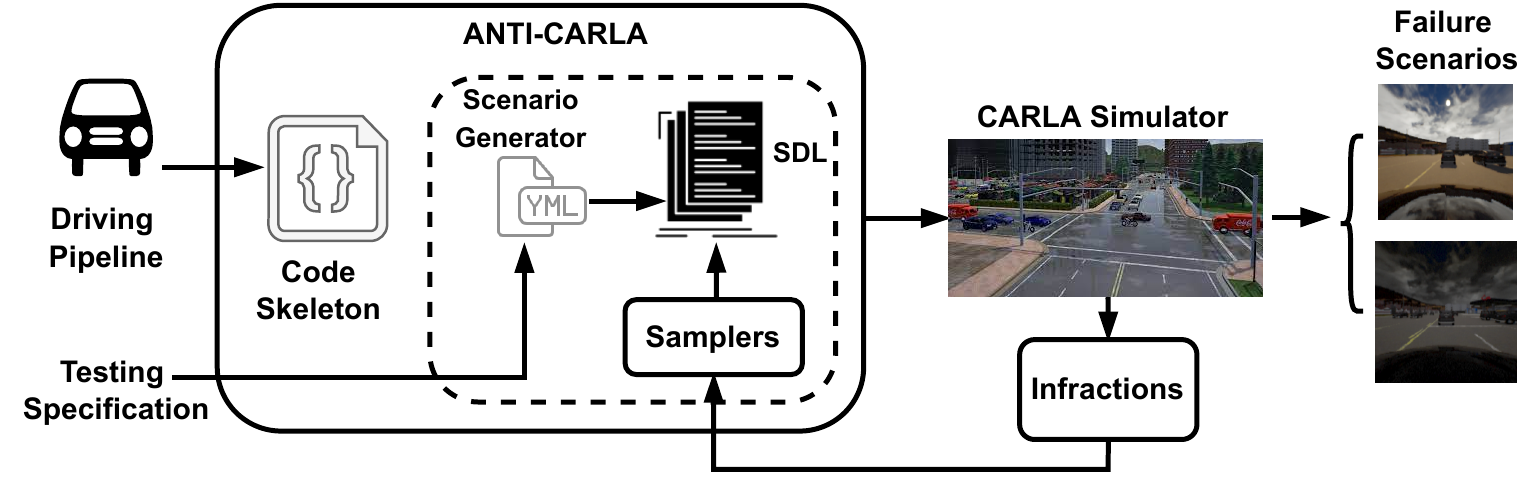}
\caption{Overview of ANTI-CARLA, the proposed adversarial testing framework integrated into the CARLA simulator. Given an arbitrary driving pipeline, the user specifies the desired test conditions, and the framework automatically generates adversarial test cases.} 
\label{fig:overview_introduction} 
\vspace{-0.1in} 
\end{figure} 

Traditionally, testing is restricted to a set of handmade scenarios with the goal of satisfying safety standards such as ISO26262~\cite{kafka2012automotive}. However, manually creating test cases is tedious and requires significant human labor. Recently, there has been substantial ongoing work in automating the test case generation process. \acp{dsml} such as Scenic \cite{fremont2019scenic}, and MSDL~\cite{msdl} are being developed for describing what a test case should contain. They provide a mechanism for describing the test conditions, the variable parameters, and their distribution, which is required for generating the test cases. These parameters are then sampled from their distribution ranges using state-of-the-art sampling algorithms such as random search, grid search, and Bayesian optimization search for generating suitable test cases. Despite this progress, the availability of frameworks that combine these components and allow straightforward test case generation is severely limited. 

Only a few frameworks have been proposed that focus on generating test cases for autonomous driving~\cite{abbas2017safe,tuncali2018sim,son2019simulation,majumdar2021paracosm}. These frameworks are built on custom simulators that are not open-source or available for use. Also, as shown in our previous work~\cite{ramakrishna2022riskaware}, the search mechanism (random and grid search) mostly used in these frameworks is inefficient in generating adversarial test cases. To address these issues, we propose ANTI-CARLA, a framework for automated adversarial testing, evaluation, and exploration of the performance of \acp{av} within the open-source CARLA simulator. It provides a skeleton that allows for plugging in and testing any autonomous driving pipeline. It includes an \ac{sdl} for describing the test conditions and a simple interface for specifying test conditions. The overall workflow of the framework is shown in \autoref{fig:overview_introduction}. Given a driving pipeline and testing specifications, ANTI-CARLA automatically generates adversarial scenarios that fail the system. Due to its flexible and modular structure, any driving pipeline can be evaluated. The main contributions of this paper are as follows.    

\begin{itemize} 
    \item We develop a domain-specific \ac{sdl} that allows for modeling different testing scenarios for the \ac{av}, defined in terms of its operating conditions. 
    
    \item We combine a mapping mechanism for an arbitrary \ac{av} system, test specification files, and an array of samplers into ANTI-CARLA to generate adversarial test cases automatically and efficiently. 

    \item We use ANTI-CARLA to evaluate the state-of-the-art \ac{lbc} controller~\cite{chen2020learning}. Despite \ac{lbc} achieving an accuracy of \SI{100}{\percent} in parts of the CARLA challenge, ANTI-CARLA generates several operating conditions that fail the controller.  

\end{itemize} 

The rest of this paper is organized as follows. In \cref{sec:rw}, we summarize related research. We discuss the problem formulation in \cref{sec:problem_formulation} and introduce the proposed framework in \cref{sec:framework}. In \cref{sec:evaluation}, we evaluate the framework by generating fail cases for the \ac{lbc} controller and analyzing them. \cref{sec:conclusion} concludes. The source code of this work will be published alongside this paper under \url{https://github.com/scope-lab-vu/ANTI-CARLA}.

%% file: individual-sections/related-works.tex
\section{Related Work}
\label{sec:rw}
In this section, we discuss related work. First, we introduce how test cases can be described and sampled. Then, we summarize existing frameworks that allow for generating test cases for a given driving pipeline. Finally, we summarize state-of-the-art driving pipelines for CARLA.

\subsection{Test Case Description and Sampling}

For testing software, test case generation has been a relevant research field for decades~\cite{rayadurgam2003generating}. The field of \ac{av} testing has only recently started gathering interest. Domain-specific \acp{sdl} have been increasingly used for specifying the testing conditions~\cite{fremont2019scenic,dreossi2019verifai,schutt2020sceml}. For example, Scenic~\cite{fremont2019scenic} is a popular language integrated with the CARLA simulator for setting up different scenarios. MSDL~\cite{msdl} is a language predominantly used in the industry to specify scenarios. Over the last years, languages such as SceML~\cite{schutt2020sceml} have been developed to enhance the capabilities during testing. 

To cover the operating conditions space, these languages are integrated with probabilistic samplers. Passive samplers such as the random and grid search are widely adopted to sample from the search space. These samplers do not use the feedback of previous results in the sampling process. For example, in grid search, a previously identified risky scenario is used as the initial condition, and the grid around it is searched to sample new scenarios~\cite{ding2020learning}. Despite their wide usage, these samplers require prior information on the interesting regions of the search space (e.g., areas of high risk to the system under test) and can be labor-intensive. An array of active samplers has been proposed in recent years, which use the feedback of previous results to make the sampling process more efficient. For example, sampling techniques such as incremental sampling, importance sampling, and adaptive sampling, borrowed from other fields such as uncertainty quantification and design space exploration~\cite{dalbey2021dakota}. Zhao~et~al.~\cite{zhao2017accelerated} used importance sampling to learn the parameters affecting the performance of the system under test, generating increasingly varied test cases that fail the system. The VERIFAI~\cite{dreossi2019verifai} software toolkit provides an array of passive samplers (e.g., random search, grid search, Halton sequence search~\cite{halton1960efficiency}) and active samplers (e.g., Bayesian optimization search) to generate test cases.

We integrate both passive and active samplers into our framework. Due to the modular design of ANTI-CARLA, existing active samplers can be integrated easily.

\begin{figure*}[t]
\centering
 \includegraphics[width=0.9\textwidth]{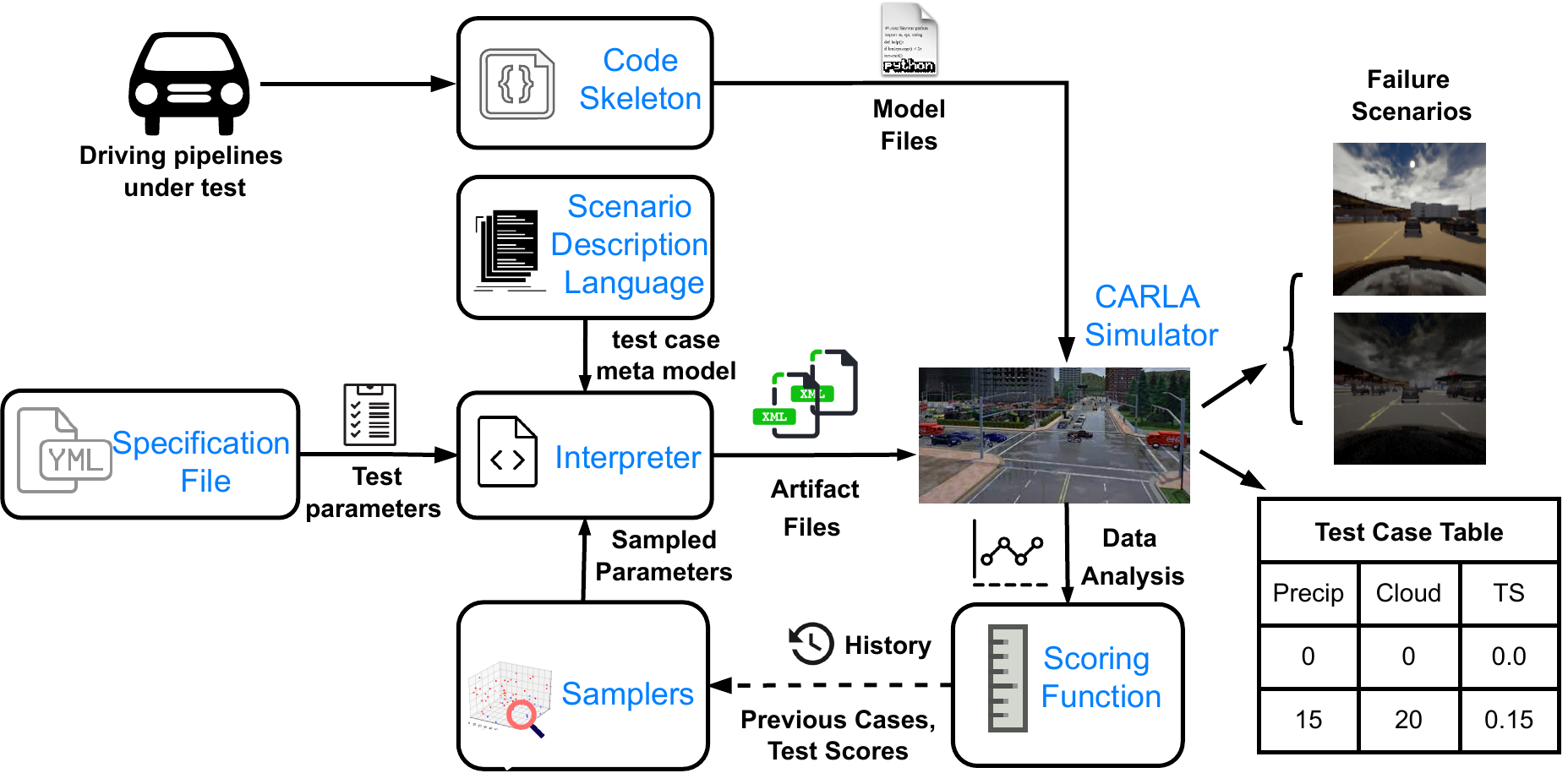}
 \caption{Overview of the ANTI-CARLA framework for generating test cases that fail an \ac{av} system in the CARLA simulator. The driving pipeline and the test specifications are taken as inputs by the framework to generate failure test cases that can be post processed to analyze the problems with the system.}
 \label{fig:workflow}
\end{figure*}


\subsection{Testing Frameworks}

Simulation-based testing frameworks allow generating test cases for a given system. A framework consists of two main components: a mechanism for describing test cases, and a mechanism for generating test cases. Tuncali~et~al.~\cite{tuncali2016utilizing} presented a testing framework built using their MATLAB tool called S-TaLiRo. The tool provides an optimization engine and a stochastic sampler to automatically sample test cases across the search space generated from the testing specification. The same authors later presented a simulation-based adversarial testing framework called Sim-ATAV~\cite{tuncali2018simulation}, which performs adversarial testing in the scenario configuration space of perception-based \ac{av}s. 
Son~et~al.~\cite{son2019simulation} proposed an \ac{adas}-testing framework based on co-simulation of Siemens Amesim and Prescan software. The testing is driven by several handmade scenarios derived from safety standards such as ISO26262. The Paracosm framework~\cite{majumdar2021paracosm}  allows users to describe complex driving situations as programs. It provides a systematic approach for exploring the search space and contains a coverage metric to quantify the coverage. For Grand Theft Auto, a simulation-based harness for \ac{av} is available~\cite{abbas2017safe}. It includes a testbench that performs sampling using simulated annealing to sample the next simulation parameters based on the \ac{av}'s performance in the current simulation. While all of these works focused on \ac{av} testing, they are implemented either in proprietary simulators or in limited custom scenarios. 

To the best of our knowledge, no framework is readily available for testing \acp{av} in open-source simulation. Despite CARLA being a widely used simulator across academia and industry, a flexible framework for testing \acp{av} is currently unavailable for CARLA. This motivates the introduction of ANTI-CARLA for adversarial testing of arbitrary driving pipelines with minimal effort.

\subsection{Autonomous Driving Pipelines}
The proposed framework is designed to generate fail cases for a given autonomous driving pipeline. A range of controllers that perform well in CARLA is available. The Transfuser approach~\cite{prakash2021multi} uses transformers to fuse LIDAR and camera input. The World-on-Rails pipeline~\cite{chen2021worldonrails} only relies on camera images. It simplifies the driving task by assuming that the vehicle does not influence the environment, then uses reinforcement learning to obtain a driving policy. Finally, the \ac{lbc} approach~\cite{chen2020learning} is also based on visual input. First, a teacher agent with complete access to internal simulator information is trained to imitate expert trajectories. Next, the actual controller is trained to imitate the teacher agent's trajectories. By learning from a teacher who was ``cheating" by having complete information, the student network is capable of learning highly accurate driving strategies. \ac{lbc} achieved an accuracy of \SI{100}{\percent} on the original CARLA benchmark~\cite{chen2020learning}. Current benchmarks are thus not always capable of identifying weaknesses of a controller, demonstrating the need for adversarial testing as offered by ANTI-CARLA. 

While any controller could be used, we select \ac{lbc} as a representative state-of-the-art approach to evaluate with ANTI-CARLA. 



%% file: individual-sections/problem-statement.tex
\section{Problem Formulation}
\label{sec:problem_formulation}

Formally, the problem can be defined as follows. We have a simulator $Sim$ that is given the current environment $E_t$, an \ac{av} with a driving controller $C$, and the current state of the \ac{av} $\textit{state}_t$. The state of the vehicle $\textit{state}_t \in \mathbb{R}^n$ describes the position, speed, and steering angle of the vehicle. The current environment $E_t$ can be defined in terms of a set of temporal variables such as weather, traffic density, the time of day that change over time and a set of spatial variables that describe roadway features. These features are characterized by waypoints $w$ denoted by a two-dimensional matrix of latitude and longitude. These waypoints can be mapped to different road segments that constitute a track on which the \ac{av} is tested. Further, the controller $C$ is a function that perceives the environment using measurements from multi-modal sensors such as a camera, LIDAR, Radar, etc. to compute actuation controls of speed, throttle, and brake. The control signals are sent to the simulator, which generates the next state of the vehicle $\textit{state}_{t+1}$ by running the controller $C$ under the given environment variables $E_t$:  $Sim(state_i,E_i,C) = state_{i+1}$. An ordered sequence of $n$ consecutive states $state_i$, $i \in \{t-n,...,t\}$ is considered a $\textit{scene}$. The simulator has some infraction function $I$ that records for each state its infractions $I(\textit{state}_t) \in \mathbb{R}_n$. A scoring function $TS$ assigns a score to the entire scene: $TS(\textit{scene}) =  \sum_{k=t-n}^{t} \textit I(state_t)$. Then, we want to solve the following problem.

\begin{problem}
\label{prob:high-risk scene}
Given a simulator $Sim$, a driving controller $C$ and some conditions on the environment $E$, generate a set of scenes that results in high infraction scores $TS$.
\end{problem}


Addressing this problem requires a framework that allows for (1) specifying the operating conditions in which the system vehicle is expected to operate. (2) finding the conditions that could result in a high infraction score, and (3) integrating a given controller $C$ into the given simulator $Sim$. In the next section, we introduce ANTI-CARLA to implement these requirements.

%% file: individual-sections/proposed-framework.tex
\section{ANTI-CARLA Framework}
\label{sec:framework}

In this section, we discuss the proposed ANTI-CARLA framework and its components shown in \cref{fig:workflow}.    


\subsection{Scenario Generator}

The first component of the framework is a scenario generator. It includes a \acl{sdl} for modeling scenarios as well as specification files for specifying and selecting testing parameters. We define a test case in terms of a scene, which is a time-series trajectory of the system's path in the environment that lasts \SI{30}{\second} to \SI{60}{\second}. A scene is represented using the environmental conditions ($E$) and the \ac{av} system's parameters ($A$). $E$ can be defined using structural features (e.g., type of road and road curvature) and temporal features (e.g., weather and traffic density). $A$ includes information like the starting position, onboard sensors, and actuators. Together, $E$ and $A$ form the testing parameter set. The value for some of these parameters can be sampled from specified distributions. 
Further, the sampling process is governed by a set of physical constraints that limit the rate at which these parameters can evolve. For example, the time of day has a fixed rate at which it can change. Including these constraints during sampling results in more meaningful scenes, as shown in our previous work~\cite{ramakrishna2022riskaware}.   

\begin{figure}[t]
\centering
 \includegraphics[width=\columnwidth]{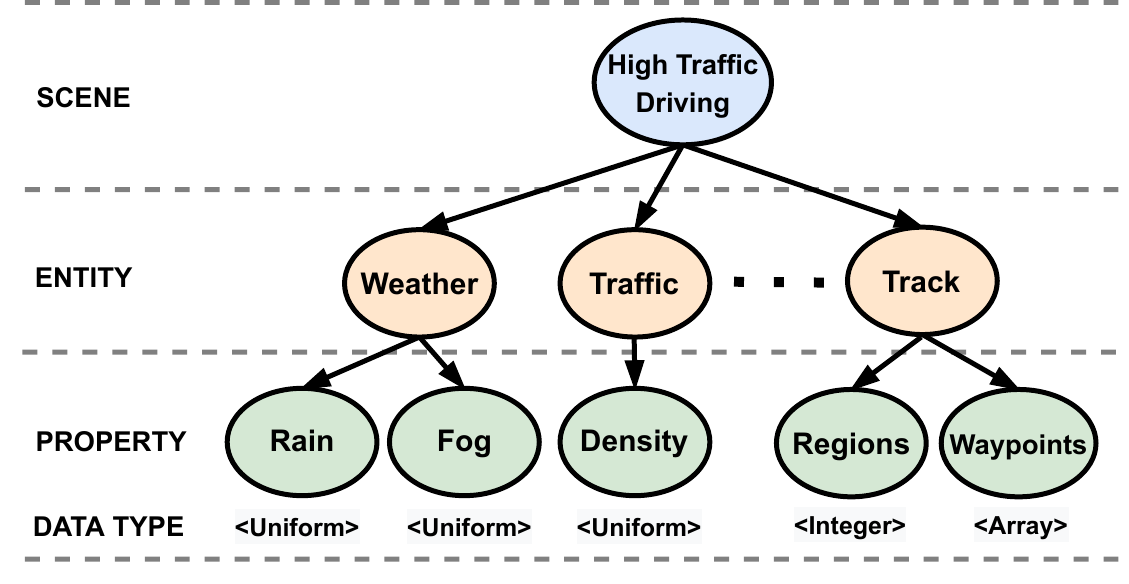}
 \caption{Example for the structure of the proposed meta-model.}
 \label{fig:meta}
\end{figure}

\textbf{\acl{sdl}}: We have designed an \ac{sdl} in the textX \cite{dejanovic2017textx} meta-language for modeling a scene. A grammar contains the rules for describing scene in the meta-language. A meta-model contains the actual description of a scene. A visualization of the structure of the meta-model is shown in \cref{fig:meta}. A scene $s$ is a collection of entities \{$e_1$,$e_2$,...,$e_k$\} that represent different environmental conditions and agent parameters. For example, the scene \textit{High Traffic Driving} is specified by its \textit{Track} as well as its \textit{Weather} and \textit{Traffic} conditions. Entities are further specified by a set of properties \{$p_1$,$p_2$,...,$p_m$\}. For example, the \textit{Weather} can have the properties of \textit{Rain} and \textit{Fog}. Each property has a name $n$ and a data type $t$. For example, \textit{Rain} is a uniform distribution, while the \textit{Waypoints} are an array of integers. Special data types such as distributions can again have properties, e.g. the range of a uniform distribution. The meta-model allows for a structured description of any desired scene in the CARLA simulator.

\textbf{Specification Files}: We also provide a set of specification files with the scene parameters that can be selected by the user. These files serve as an abstract representation of the \ac{sdl} modeling concepts. Based on the user's selection of the sampler, the selected parameters are sampled from their respective distributions. The remaining parameters are assigned default values. We divide the information into three specification files. First, the user selects a scene specifier. We show an excerpt of the scene specifier file in \autoref{fig:scene-specification}. It specifies which town to use as a map, which track to drive, the distributions for the weather traffic density, pedestrian density, and the sampling constraints. The user also specifies which infraction metric to use and at which frequency data should be recorded. Second, an agent specifier is needed, which includes agent-related information such as the available controllers, list of sensors and their positions, and the sensor data that can be recorded. Third, a sampler specifier determines which sampler to use from a list of available samplers. 

Finally, the language has an interpreter as shown in \cref{fig:workflow}. It connects the specification files, the \ac{sdl}, and the probabilistic samplers discussed in the following section. First, the interpreter extracts the parameters that require sampling and the fixed parameters from the specification files. It then sends the parameters that need sampling to the sampler. The sampler generates a value for these parameters from their distribution ranges. Finally, the sampled parameters and the fixed parameters are parsed to the \ac{sdl} meta-model to generate artifact files that drive the simulator. 

\input{specification-file}

\subsection{Adapter Glue Code}
\begin{figure}[t]
\centering
 \includegraphics[width=\columnwidth]{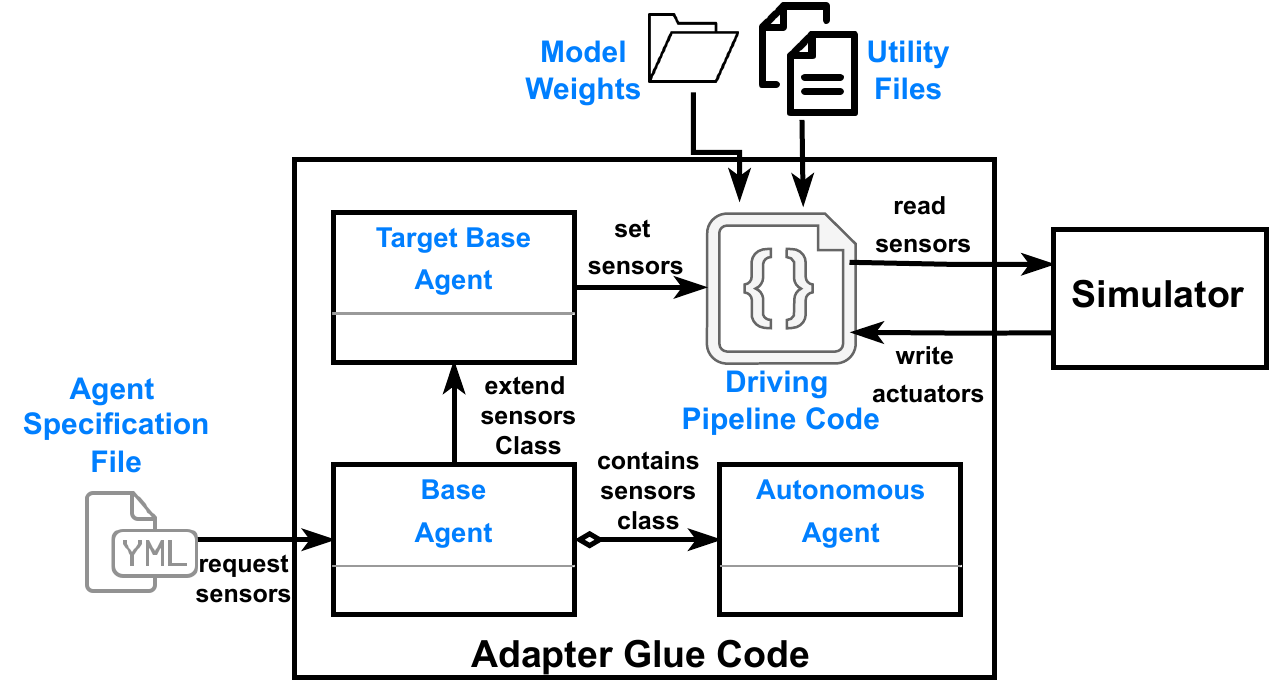}
 \caption{An illustration of the adapter glue code that interfaces the sensors and actuators of CARLA to the code of the \ac{av} system.}
 \label{fig:skeleton}
\vspace{-0.1in}
\end{figure}

The framework also requires a driving pipeline under test. Integrating different pipelines is not straightforward since they might not have the right interface to be used ``as in" the framework. For example, driving pipelines developed outside of CARLA may not be directly used in the simulator because of strict interface requirements. They may need to be ``adapted" to meet the interface expectations of the simulator~\cite{adapter}. To address this, we generate an adapter that interfaces the driving pipeline code with the sensors and actuators in the format required by the simulator's API as shown in \autoref{fig:skeleton}. The adapter is synthesized from the agent specification file, which has a list of sensors required by the driving pipeline, its positions, and the sampling rates. 

The adapter reads the available sensors from the autonomous agent class in the simulator's API and extends it with the sensors requested in the specification file. Thus, a code structure for the requested sensors is generated and provided to the driving pipeline code. Also, the actuators required by the simulator are read from the API and made available in the code. With the required sensors and actuators available through the adapter, the user needs to provide the driving pipeline code. There are specific directories for any utility files and model weights of the controller, which are linked with the code skeleton. The user only needs to handle this interface for setting up their controller correctly. If the actuators and sensors are not properly set up, the simulator will throw an error. In future work, we will include an automatic check for ensuring a correct-by-construction setup between the code, sensors, and actuators.

\subsection{Scoring Function}
For evaluating the driving proficiency of the \ac{av} system in the generated scene, the framework also provides a scoring function called test score $TS$. The scoring is performed based on all of the infractions performed by the system in a given scene. Infractions measured in CARLA include route deviation, lane violation, traffic rule violation, running a stop sign, running a red light, and off-road driving. Each infraction $I_k$ is assigned a weight $w_k$. In the current setup, these weights are set to the values used in the CARLA challenge~\cite{carla-challenge}. However, they can be varied depending on the current use case. The infractions are then combined into the weighted score as shown in \autoref{eqn:driving_score}:

\begin{equation}
    \textit{TS} = \sum_{k=1}^{n} w_k \cdot I_k
\label{eqn:driving_score}
\end{equation}

The test score $TS$ generated from the infractions is stored along with the test case parameters in a test case table as shown in \autoref{fig:workflow}. These test cases and the scores are used online to drive the active samplers towards regions of the search space that have previously resulted in high test scores. The table can also be used to perform an offline post-analysis to identify the failure conditions of a given controller, allowing to retrain and improve the controller.

\subsection{Samplers}
We have integrated several samplers to perform search-based test case generation. A sampler is interfaced to the \ac{sdl} through an interpreter, which provides it with the scene parameters and the distribution from which different values for the parameters can be sampled.

We included two kinds of samplers available in the framework. First, we implemented several passive samplers since they are fast and widely used. They do not use the feedback of previous results in the sampling process. \textit{Random Search}, uniformly samples the parameter value from their respective distributions at random. \textit{Grid Search} exhaustively searches all of the combinations of the parameters in a given grid. \textit{Halton Sequence Search}~\cite{halton1960efficiency} is a pseudo-random technique that samples the parameters using co-primes as their bases. While these samplers perform well, their non-feedback sampling approach results in a directionless search that could miss several important failure test cases. Further, they do not balance the exploration vs. exploitation of the search space, which is required for generating diverse failure cases~\cite{ramakrishna2022riskaware}.

To overcome these limitations, we have also included two adversarial samplers, \textit{\ac{rns}} and \textit{\ac{gbo}} that we developed in our previous work~\cite{ramakrishna2022riskaware}. These samplers use the feedback of the system's previous performances when sampling the parameters for the current test case. In addition, they also capture constraints and correlations between the different test parameters to generate meaningful test cases. The overall idea with the feedback and the constraints is to move the sampling process towards the regions of the search space that are highly likely to fail the system. 

The \textit{\ac{rns}} sampler extends the conventional random search with the kd-tree nearest neighborhood search algorithm.
This extension provides the random sampler with the capability to exploit. If the test case generated from randomly sampled parameters results in a high test score, the region around these parameter values is exploited. Otherwise, the parameters are again randomly sampled from the entire distribution. The \textit{\ac{gbo}} extends the conventional \acl{bo} sampler with constraints, which restrict the region where the acquisition function looks for the next sampling variables.

%% file: specification-file.tex
\begin{figure}[!t]
\setlength{\abovecaptionskip}{-3pt}
\begin{lstlisting}[language=specification,numbers=none]
Scenario Description{
    town: 5 //Available towns 3 and 5
    track: 1 // 1 track available for each town
       regions: 5 //Each town has 5 regions
    weather:  //Weather parameters and distribution range
      cloudiness: [0,100]  
      precipitation: [0,100] 
      time-of-day: [-90,90] 
    pedestrian_density: [0,3] 
    traffic_density: [0,10]
    Constraints: //A constraint on the rate of change in parameter values
        weather_delta: 2
        traffic_delta: 2
        pedestrian_delta: 1
    Infraction_Metrics: //Infraction metrics to be recorded
        Infraction Penalty: true
        Off-road Driving: true
        Route Deviation: false
    Record Frequency: 5Hz } //Frequency of data recording 

\end{lstlisting}
\caption{Excerpt of a scene specification file. The specification files describe the available inputs, the user then only has to select a value for each concept.}
\label{fig:scene-specification}
\end{figure}


%% file: individual-sections/evaluation.tex
\section{Evaluation}
\label{sec:evaluation}

In this section, we present several experiments to evaluate the usability of ANTI-CARLA for adversarial testing. We use the framework to compare the adversarial test cases generated by different samplers. Then, we compare the performance of the \ac{lbc} controller to the Transfuser approach~\cite{prakash2021multi} on those test cases. The experiments were run on a desktop computer with AMD Ryzen Threadripper 16-Core Processor, 4 NVIDIA Titan XP GPUs, and 128 GiB RAM. 

\begin{figure}[t!]
\centering
 \includegraphics[width=\columnwidth]{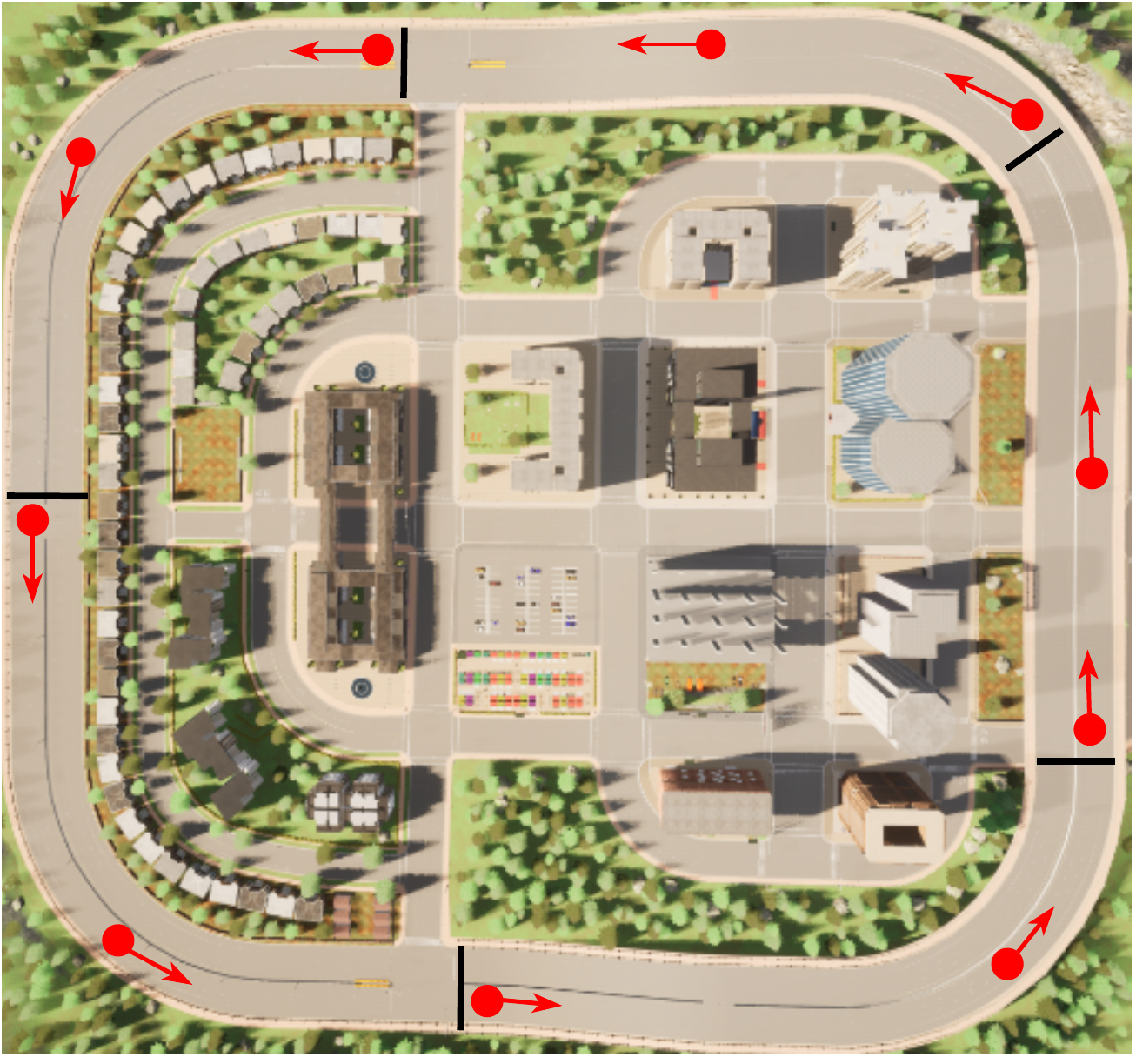}
 \caption{An illustration of the track in CARLA's town5. Red arrows highlight the waypoints and the five regions are separated by black lines.}
 \label{fig:track}
\end{figure}

\begin{figure*}[t!]
 \centering
 \includegraphics[width=\textwidth]{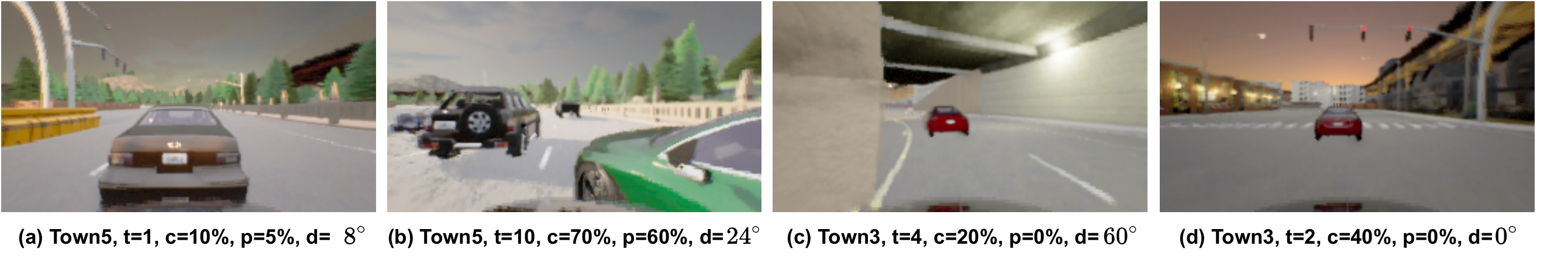}
 \caption{Screenshots of the test cases as captured by the forward-looking camera of the \ac{av}. Descriptions of these scenes are provided below the images.}
 \label{fig:scenes}
\end{figure*}

\subsection{Simulation Setup}
The proposed framework is integrated into the CARLA simulator. We used the CARLA challenge API~\cite{carla-challenge} to create one closed loop track in both towns 3 and 5. We use \ac{lbc} as the controller $C$ under test. The geometry of the track is defined by ten waypoints as shown in \autoref{fig:track}. We divide the track into several regions containing two waypoints each. For each track, we thus obtain five regions. We divided each track into regions to create shorter scenes in which we can vary the weather conditions as well as traffic and pedestrian densities. Each track can then have ten different environmental conditions. The length of the track and number of regions can be specified in the scene description file and can thus be changed with minimal effort in the code. We use the API to create and control the traffic and pedestrians in each scene. For evaluating each simulation, we record the driving score, the route completion score, and the test score $TS$ based on the list of infractions available from the API.

\subsubsection{Comparison Metrics}
Besides evaluating the performance of the driving pipeline, the framework also allows us to evaluate the performance of the samplers used to generate adversarial test cases. We use the following two metrics. 

\textbf{Failed Test Cases (FT)}: This score measures the efficacy of the samplers in generating failure test cases. It is calculated as the number of failed test cases $N_{Fail}$ compared to the total number of sampled test cases $N_{Total}$:

\begin{equation}
\label{eqn:ps}
\small
    FT (\%) = \frac{N_{Fail}}{N_{Total}} \cdot 100
\end{equation}

\textbf{Total Execution Time}: The overall time taken by the sampler to sample $N_{Total}$ test cases and execute them in the simulator is a metric relevant for practice.

\subsection{Results}
We generated $100$ test cases each for the track in town3 and town5. Each test case represents a scene that lasts between $30$ seconds to $60$ seconds. We varied the environment parameters of cloudiness $c$, precipitation $p$, time of day $d$, and traffic density $t$ to generate different test cases. We varied the cloudiness and precipitation in the range $[0,100]$, the time of day in the range $[0,90]$, and the traffic in the range $[0,50]$. We used the initial conditions of $d$ = $0^{\circ}$ (dusk), $c$ = $0^{\circ}$, $p$ = $0^{\circ}$, and $t$ = 5. To score the test cases, we computed a test score as $TS = R_i \cdot I_S$. Here, $R_i$ is the route completion percentage of the $i^{th}$ route, and $I_S$ is the weighted sum of major infractions such as collision with pedestrians, collisions with vehicles, collisions with static objects, and minor infractions such as running a stop sign, running a red-light signal, and off-road driving. The weights for these infractions are taken directly from the CARLA challenge setup. To drive the test case generation process, we selected the commonly used random sampler as a baseline and compared it against the \ac{rns} sampler. The simulator was run in synchronous mode at a fixed rate of $20$ frames per second. If a test case includes either a collision, infraction, or route in-completion, it is considered to be a failure test case. 

\subsubsection{Visualization}
First, we visualize several test cases by showing the frontal camera view from each scene. In \cref{fig:scenes}, we show four exemplary test cases for \ac{lbc}. \cref{fig:scenes}-a is a nominal test case from town5 with low precipitation and dusk time of the day and low traffic. \cref{fig:scenes}-b is a failed test case from town5 with high precipitation and traffic. A collision occurred when the controller tried to steer to the right lane. \cref{fig:scenes}-c is a failed test case from town3 with no precipitation and low traffic. Here, the \ac{av} can be seen navigating a tunnel. Towards the end of the tunnel, the \ac{av} collides with a pillar. Finally, \cref{fig:scenes}-d is another failed test case from town3, where the \ac{av} runs a red traffic light. These examples demonstrate that a diverse set of fail cases could be obtained.

\input{results/town_stats}

\subsubsection{Sampler Comparison}

\cref{Table:stats} shows the statistics of the collisions and infractions generated by the two samplers across all test cases. The \ac{rns} sampler generates a higher number of failed test cases than the random sampler. In general, the \ac{av} performed better in town5 than in town3. The track in town5 was shorter, had fewer traffic lights and stop signs, and did not have complex landmarks such as a tunnel or a round-about. The controller failed \SI{13}{\percent} and \SI{21}{\percent} of the test cases generated by the random and \ac{rns} sampler, respectively. Town3 has several traffic lights and a tunnel. Here, the controller failed \SI{17}{\percent} and \SI{27}{\percent} of the test cases generated from the random and \ac{rns} sampler, respectively. These fail cases occurred in the region that included the tunnel as well as areas with traffic signs and stop signs nearby. The framework required $140$ minutes to generate the test cases for town5 and $325$ minutes for town3 when using the random sampler. With the \ac{rns} sampler, the execution times are $176$ minutes for town5 and $384$ minutes for town3. This shows that the more efficient \ac{rns} sampler does not add significant overhead.

\begin{figure}[t]
 \centering
 \includegraphics[width=\columnwidth]{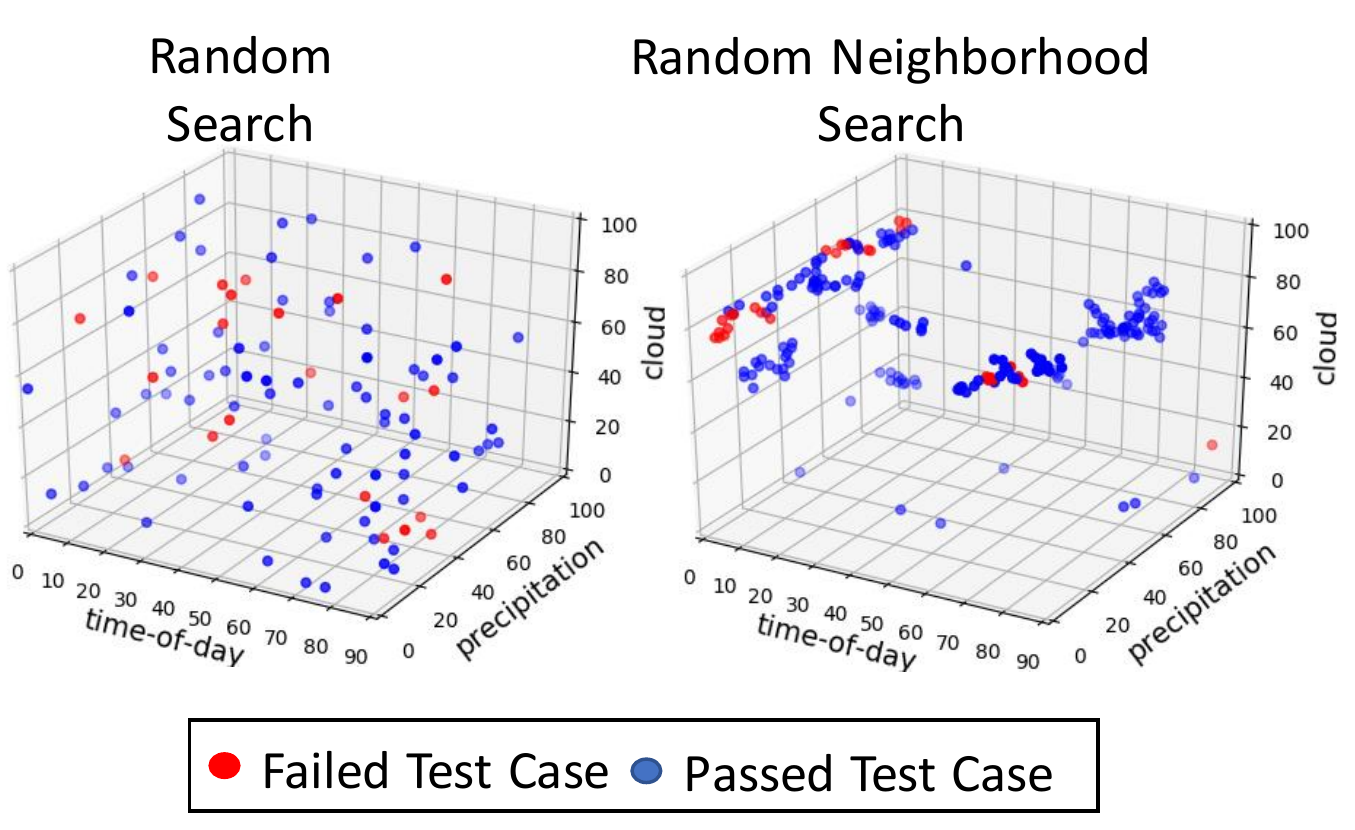}
 \caption{Comparison of the $100$ scenes sampled by the random and \ac{rns} samplers. The \ac{rns} sampler generates distinct clusters of failures. Plot axis: x-axis represents the time of day, y-axis represents the precipitation level and z-axis represents the cloud level.}
 \label{fig:sampler_3d}
\end{figure}

\cref{fig:sampler_3d} shows the test cases sampled by the random and \ac{rns} sampler plotted in the operating conditions space. The failed test cases are marked in red, and the test cases that passed without failures or collisions are marked in blue. The random sampler randomly samples the test cases across the search space. This makes it hard to analyze common causes of the controller's failures. In contrast, the failed test cases generated by the \ac{rns} sampler occur in clusters, which makes it easier to hypothesize the causes of the failures. The figure shows that the controller had failures mostly in high precipitation and dusk (time of the day) operating conditions. 

\subsubsection{Town-wise Infractions}

\cref{fig:performance} shows the infraction statistics of the \ac{lbc} controller gathered from the $100$ test cases generated by the \ac{rns} sampler across towns 3 and 5. $I_C$ includes collisions with vehicles, pedestrians, and static obstacles. Other infractions are running a red light $I_R$, running a stop sign $I_S$, route deviations $I_{RD}$, off lane driving $I_{OL}$, and simulation timeout $I_T$. A timeout occurs if the vehicle is involved in an unrecoverable collision or if the controller operating the vehicle unexpectedly stops with a hard brake signal. The controller had a total of $27$ failure test cases in town3. Among these, the vehicle was involved in a collision in $13$ test cases, red-light violations in $7$, stop-sign violation in $5$, a route deviation in $1$, off-lane driving in $1$, and a timeout in $2$ test cases. In comparison, the vehicle had fewer failure test cases in town 5. The vehicle had fewer collisions, red-light violations, and stop sign violations. However, there were more test cases in which the vehicle had route deviations and off-lane driving.

\input{results/town_comparison}

\subsubsection{Controller Comparison}
ANTI-CARLA can also be used to compare different controllers. We compare the performance of the Transfuser~\cite{prakash2021multi} controller to \ac{lbc} using the track in town5. We ran the Transfuser and \ac{lbc} for the same $100$ test cases generated with the random sampler. The \ac{lbc} and Transfuser controllers had $13$ and $23$ failed test cases, respectively. \cref{fig:controller-comparison} shows a breakdown of the infractions caused by these controllers. Simulation timeout is the main reason for the Transfuser's higher failure rate. This could be due to the expert policy used during training not being sufficient for handling all scenarios. Besides the frequent timeouts, the Transfuser had significantly fewer collisions and infractions than \ac{lbc}. Both controllers struggled with detecting red traffic lights. The authors of Transfuser suggested that this is due to traffic lights being placed on the opposite side of intersections, which is difficult to detect in camera images~\cite{prakash2021multi}. These results show how ANTI-CARLA allows us to identify the weaknesses of different controllers by focusing specifically on adversarial test scenarios. 


\input{results/controller-comparison}

\subsubsection{Recommendations}
The previous sections show that \ac{lbc} tends to crash into the leading vehicle in heavy rain or in unusual lighting conditions, caused either by the time of day or by entering a tunnel. The highest infraction scores are obtained for adverse weather conditions and challenging landscapes such as traffic lights, tunnels, and roundabouts. By identifying those main reasons for failures, several suggestions can be made for improving \ac{lbc}. First, adding another sensor modality could make the perception more robust to rain or darkness. For example, LIDARs are less susceptible to scene lighting levels or rain intensity. Second, the robustness of the camera-based perception could be improved by training the controller with more images taken in the adverse conditions of rain and at dusk.  

%% file: results/town_stats.tex
\begin{table}[t]
\centering
\renewcommand{\arraystretch}{1.3}
\footnotesize
\begin{tabular}{|c|c|c|c|c|}
\hline
\textbf{Samplers}                & \textbf{Town} & \textbf{\begin{tabular}[c]{@{}c@{}}Failed \\ Test \\ Cases (\%)\end{tabular}} & \textbf{\begin{tabular}[c]{@{}c@{}}Test cases \\ with \\ collisions (\%)\end{tabular}} & \textbf{\begin{tabular}[c]{@{}c@{}}Test cases \\ with \\ infractions\\ (\%)\end{tabular}} \\ \hline
\multirow{2}{*}{\textbf{Random}} & 3             & 17                                                                            & 9                                                                                      & 10                                                                                     \\ \cline{2-5} 
                                 & 5             & 13                                                                            & 7                                                                                      & 8                                                                                      \\ \hline
\multirow{2}{*}{\textbf{RNS}}    & 3             & 27                                                                            & 13                                                                                     & 15                                                                                     \\ \cline{2-5} 
                                 & 5             & 21                                                                            & 10                                                                                      & 13                                                                                     \\ \hline
\end{tabular}
\caption{Test case statistics for the random and \ac{rns} sampler.}
\label{Table:stats}
\vspace{-0.1in}
\end{table}

%% file: results/town_comparison.tex
\begin{figure}[t]
\centering
\begin{tikzpicture}
\footnotesize

\begin{groupplot}[group style = {group size = 1 by 1, horizontal sep = 55pt}, width = 6.0cm, height = 5.0cm]
        \nextgroupplot[ 
            height=4.5cm, width=7.5cm,
        bar width=0.1cm,
    ybar,
    enlargelimits=0.15,
    x grid style={white!69.0196078431373!black},
xmajorgrids,
y grid style={white!69.0196078431373!black},
ymajorgrids,
    legend style={at={(0.7,0.9),font=\footnotesize},
      anchor=north,legend columns=-1},
    ylabel style={align=center}, ylabel= \# of Infractions,
    symbolic x coords={$I_C$,$I_R$,$I_S$,$I_{RD}$,$I_{OL}$,$I_T$},
    xtick=data,
    xlabel=Type of Infractions,
    xtick pos=left,
ytick pos=left,
    nodes near coords align={vertical},
    ]
\addplot [red!20!black,fill=red!80!white]coordinates {($I_C$,13)($I_R$,8)($I_S$,3)($I_{RD}$,1)($I_{OL}$,1)($I_T$,2)};
\addplot coordinates {($I_C$,10)($I_R$,7)($I_S$,0)($I_{RD}$,2)($I_{OL}$,2)($I_T$,2)};

        
\legend{Town3,Town5}             
    \end{groupplot}

\end{tikzpicture}
\caption{Town-wise infractions of the \ac{lbc} controller. The infractions included in the plot are: $I_C$ includes collisions with vehicles, pedestrians, and static obstacles. Other infractions are running a red light $I_R$, running a stop sign $I_S$, route deviations $I_{RD}$, off lane driving $I_{OL}$, and simulation timeout $I_T$.}
\label{fig:performance}
\vspace{-0.1in}
\end{figure}

%% file: results/controller-comparison.tex
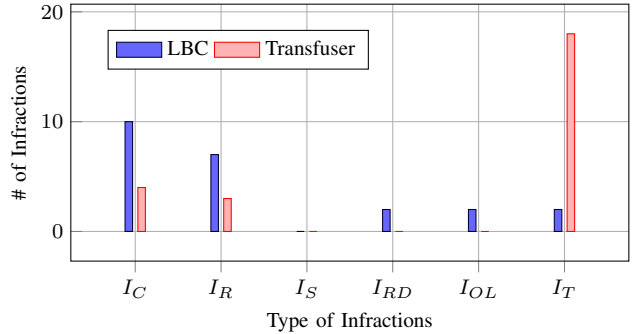
\begin{figure}[t]
\centering
\begin{tikzpicture}
\footnotesize

\begin{groupplot}[group style = {group size = 1 by 1, horizontal sep = 55pt}, width = 8.0cm, height = 8.0cm]
        \nextgroupplot[ 
            height=5cm, width=9cm,
        bar width=0.1cm,
    ybar,area legend,
    enlargelimits=0.15,
    x grid style={white!69.0196078431373!black},
xmajorgrids,
y grid style={white!69.0196078431373!black},
ymajorgrids,
    legend style={at={(0.3,0.9),font=\footnotesize},
      anchor=north,legend columns=-1},
    ylabel style={align=center}, ylabel= \# of Infractions,
    symbolic x coords={$I_C$,$I_R$,$I_S$,$I_{RD}$,$I_{OL}$,$I_T$},
    xtick=data,
    xlabel=Type of Infractions,
    xtick pos=left,
ytick pos=left,
    nodes near coords align={vertical},
    ]
\addplot [blue!20!black,fill=blue!60!white]coordinates {($I_C$,10)($I_R$,7)($I_S$,0)($I_{RD}$,2)($I_{OL}$,2)($I_T$,2)};
\addplot coordinates {($I_C$,4)($I_R$,3)($I_S$,0)($I_{RD}$,0)($I_{OL}$,0)($I_T$,18)};

        
\legend{\ac{lbc},Transfuser}             
    \end{groupplot}

\end{tikzpicture}
\caption{Infraction breakdown caused by the \ac{lbc} and Transfuser controllers. The infractions included in the plot are: $I_C$ includes collisions with vehicles, pedestrians, and static obstacles. Other infractions are running a red light $I_R$, running a stop sign $I_S$, route deviations $I_{RD}$, off lane driving $I_{OL}$, and simulation timeout $I_T$.}
\label{fig:controller-comparison}
\vspace{-0.1in}
\end{figure}

%% file: individual-sections/conclusion.tex
\section{Conclusion} 
\label{sec:conclusion} 
In this paper, we introduced the ANTI-CARLA framework for the CARLA simulator that allows to automatically generate test cases that fail an arbitrary \ac{av} system. Testing frameworks available in the literature are either tailor-made for a specific use case or built on proprietary simulators. In contrast, ANTI-CARLA is an open-source extension to an open-source simulator. The framework integrates a \acl{sdl} for modeling test scenarios in terms of the system's operating conditions, sensor, and actuator faults. A test specification file is used to specify and select the test conditions, infraction metrics, and samplers required for generating the test cases. The \ac{sdl} is driven by an adversarial sampler that searches across the specified operating conditions space. We used this framework to test the popular \acl{lbc} controller and to compare different samplers. We also used ANTI-CARLA to compare the performance of \ac{lbc} to the Transfuser approach, allowing us to identify the weaknesses of each controller. 


We plan to move this research in several directions. First, currently, only static scenes can be sampled. A temporal sequence of scenes leading up to each fail case is unavailable. Temporal and dynamic sampling process using \acl{rl} and Monte Carlo tree search will be added in the future. Second, the sampling process is currently slow, with each test case taking approximately $5$ minutes. To scale the testing process, we will parallelize the simulations and the sampling process across dockers. Fourth, the adversarial test cases could be used for a closed-loop controller training workflow as proposed in~\cite{hartsell2019cps}. By training the controller with the obtained failure scenes and then sampling new failure scenes, the controller can be iteratively optimized.


\textbf{Acknowledgment}: This work was supported by the DARPA Assured Autonomy project and Air Force Research Laboratory. Any opinions, findings, and conclusions expressed in this material are those of the author(s) and do not necessarily reflect the views of DARPA or AFRL.

%% file: main.bbl
\begin{thebibliography}{10}
\providecommand{\url}[1]{#1}
\csname url@samestyle\endcsname
\providecommand{\newblock}{\relax}
\providecommand{\bibinfo}[2]{#2}
\providecommand{\BIBentrySTDinterwordspacing}{\spaceskip=0pt\relax}
\providecommand{\BIBentryALTinterwordstretchfactor}{4}
\providecommand{\BIBentryALTinterwordspacing}{\spaceskip=\fontdimen2\font plus
\BIBentryALTinterwordstretchfactor\fontdimen3\font minus
  \fontdimen4\font\relax}
\providecommand{\BIBforeignlanguage}[2]{{%
\expandafter\ifx\csname l@#1\endcsname\relax
\typeout{** WARNING: IEEEtran.bst: No hyphenation pattern has been}%
\typeout{** loaded for the language `#1'. Using the pattern for}%
\typeout{** the default language instead.}%
\else
\language=\csname l@#1\endcsname
\fi
#2}}
\providecommand{\BIBdecl}{\relax}
\BIBdecl

\bibitem{gibbs2017google}
S.~Gibbs, ``Google sibling waymo launches fully autonomous ride-hailing
  service,'' \emph{The Guardian}, vol.~7, 2017.

\bibitem{vlasic2016self}
B.~Vlasic and N.~E. Boudette, ``'self-driving tesla was involved in fatal
  crash,'us says,'' \emph{New York Times}, vol. 302016, 2016.

\bibitem{kohli2019enabling}
P.~Kohli and A.~Chadha, ``Enabling pedestrian safety using computer vision
  techniques: A case study of the 2018 uber inc. self-driving car crash,'' in
  \emph{Future of Information and Communication Conference}.\hskip 1em plus
  0.5em minus 0.4em\relax Springer, 2019, pp. 261--279.

\bibitem{hartsell2021resonate}
C.~Hartsell, S.~Ramakrishna, A.~Dubey \emph{et~al.}, ``Resonate: A runtime risk
  assessment framework for autonomous systems,'' in \emph{International
  Symposium on Software Engineering for Adaptive and Self-Managing Systems
  (SEAMS)}, may 2021.

\bibitem{ramakrishna2022riskaware}
S.~Ramakrishna, B.~Luo, Y.~Barve \emph{et~al.}, ``Risk-aware scene sampling for
  dynamic assurance of autonomous systems,'' \emph{arXiv preprint
  arXiv:2202.13510}, 2022.

\bibitem{habermayr2021situation}
L.~Habermayr, M.~Hofbauer, J.-V. Zacchi \emph{et~al.}, ``Situation-aware model
  refinement for semantic image segmentation,'' in \emph{2021 IEEE
  International Intelligent Transportation Systems Conference (ITSC)}.\hskip
  1em plus 0.5em minus 0.4em\relax IEEE, 2021, pp. 2696--2702.

\bibitem{dosovitskiy2017carla}
A.~Dosovitskiy, G.~Ros, F.~Codevilla \emph{et~al.}, ``Carla: An open urban
  driving simulator,'' \emph{arXiv:1711.03938}, 2017.

\bibitem{shah2018airsim}
S.~Shah, D.~Dey, C.~Lovett \emph{et~al.}, ``Airsim: High-fidelity visual and
  physical simulation for autonomous vehicles,'' in \emph{Field and service
  robotics}.\hskip 1em plus 0.5em minus 0.4em\relax Springer, 2018, pp.
  621--635.

\bibitem{rong2020lgsvl}
G.~Rong, B.~H. Shin, H.~Tabatabaee \emph{et~al.}, ``Lgsvl simulator: A high
  fidelity simulator for autonomous driving,'' in \emph{23rd International
  conference on intelligent transportation systems}.\hskip 1em plus 0.5em minus
  0.4em\relax IEEE, 2020, pp. 1--6.

\bibitem{deepdrive}
\BIBentryALTinterwordspacing
Voyage, ``Deepdrive simulator,'' 2020. [Online]. Available:
  \url{https://deepdrive.io/}
\BIBentrySTDinterwordspacing

\bibitem{hofbauer2020telecarla}
M.~Hofbauer, C.~B. Kuhn, G.~Petrovic \emph{et~al.}, ``Telecarla: An open source
  extension of the carla simulator for teleoperated driving research using
  off-the-shelf components,'' in \emph{2020 IEEE Intelligent Vehicles Symposium
  (IV)}.\hskip 1em plus 0.5em minus 0.4em\relax IEEE, 2020, pp. 335--340.

\bibitem{kuhn2021pixel}
C.~B. Kuhn, M.~Hofbauer, Z.~Xu \emph{et~al.}, ``Pixel-wise failure prediction
  for semantic video segmentation,'' in \emph{2021 IEEE International
  Conference on Image Processing (ICIP)}.\hskip 1em plus 0.5em minus
  0.4em\relax IEEE, 2021.

\bibitem{hofbauer2020multi}
M.~Hofbauer, C.~B. Kuhn, J.~Meng \emph{et~al.}, ``Multi-view region of interest
  prediction for autonomous driving using semi-supervised labeling,'' in
  \emph{2020 IEEE 23rd International Conference on Intelligent Transportation
  Systems (ITSC)}.\hskip 1em plus 0.5em minus 0.4em\relax IEEE, 2020, pp. 1--6.

\bibitem{kafka2012automotive}
P.~Kafka, ``The automotive standard iso 26262, the innovative driver for
  enhanced safety assessment \& technology for motor cars,'' \emph{Procedia
  Engineering}, vol.~45, pp. 2--10, 2012.

\bibitem{fremont2019scenic}
D.~J. Fremont, T.~Dreossi, S.~Ghosh \emph{et~al.}, ``Scenic: A language for
  scenario specification and scene generation,'' in \emph{Proceedings of the
  40th annual ACM SIGPLAN conference on Programming Language Design and
  Implementation (PLDI)}, June 2019.

\bibitem{msdl}
\BIBentryALTinterwordspacing
O.~foretellix, ``Open m-sdl.'' [Online]. Available:
  \url{https://www.foretellix.com/open-language/}
\BIBentrySTDinterwordspacing

\bibitem{abbas2017safe}
H.~Abbas, M.~O’Kelly, A.~Rodionova \emph{et~al.}, ``Safe at any speed: A
  simulation-based test harness for autonomous vehicles,'' in
  \emph{International Workshop on Design, Modeling, and Evaluation of Cyber
  Physical Systems}.\hskip 1em plus 0.5em minus 0.4em\relax Springer, 2017, pp.
  94--106.

\bibitem{tuncali2018sim}
C.~E. Tuncali, G.~Fainekos, H.~Ito \emph{et~al.}, ``Sim-atav: Simulation-based
  adversarial testing framework for autonomous vehicles,'' in \emph{Proceedings
  of the 21st International Conference on Hybrid Systems: Computation and
  Control}, 2018, pp. 283--284.

\bibitem{son2019simulation}
T.~D. Son, A.~Bhave, and H.~Van~der Auweraer, ``Simulation-based testing
  framework for autonomous driving development,'' in \emph{IEEE International
  Conference on Mechatronics (ICM)}, vol.~1.\hskip 1em plus 0.5em minus
  0.4em\relax IEEE, 2019.

\bibitem{majumdar2021paracosm}
R.~Majumdar, A.~Mathur, M.~Pirron \emph{et~al.}, ``Paracosm: A test framework
  for autonomous driving simulations,'' in \emph{International Conference on
  Fundamental Approaches to Software Engineering}.\hskip 1em plus 0.5em minus
  0.4em\relax Springer, Cham, 2021, pp. 172--195.

\bibitem{chen2020learning}
D.~Chen, B.~Zhou, V.~Koltun \emph{et~al.}, ``Learning by cheating,'' in
  \emph{Conference on Robot Learning}.\hskip 1em plus 0.5em minus 0.4em\relax
  PMLR, 2020, pp. 66--75.

\bibitem{rayadurgam2003generating}
S.~Rayadurgam and M.~Heimdahl, ``Generating mc/dc adequate test sequences
  through model checking,'' in \emph{28th Annual NASA Goddard Software
  Engineering Workshop, 2003. Proceedings.}, 2003, pp. 91--96.

\bibitem{dreossi2019verifai}
T.~Dreossi, D.~J. Fremont, S.~Ghosh \emph{et~al.}, ``Verifai: A toolkit for the
  formal design and analysis of artificial intelligence-based systems,'' in
  \emph{International Conference on Computer Aided Verification}.\hskip 1em
  plus 0.5em minus 0.4em\relax Springer, 2019, pp. 432--442.

\bibitem{schutt2020sceml}
B.~Sch{\"u}tt, T.~Braun, S.~Otten \emph{et~al.}, ``Sceml: A graphical modeling
  framework for scenario-based testing of autonomous vehicles,'' in \emph{23rd
  ACM/IEEE International Conference on Model Driven Engineering Languages and
  Systems}, 2020, pp. 114--120.

\bibitem{ding2020learning}
W.~Ding, B.~Chen, M.~Xu \emph{et~al.}, ``Learning to collide: An adaptive
  safety-critical scenarios generating method,'' in \emph{2020 IEEE/RSJ
  International Conference on Intelligent Robots and Systems (IROS)}.\hskip 1em
  plus 0.5em minus 0.4em\relax IEEE, 2020, pp. 2243--2250.

\bibitem{dalbey2021dakota}
K.~Dalbey, M.~Eldred, G.~Geraci \emph{et~al.}, ``Dakota a multilevel parallel
  object-oriented framework for design optimization parameter estimation
  uncertainty quantification and sensitivity analysis: Version 6.14 theory
  manual.'' Sandia National Lab.(SNL-NM), Albuquerque, NM (United States),
  Tech. Rep., 2021.

\bibitem{zhao2017accelerated}
D.~Zhao, X.~Huang, H.~Peng \emph{et~al.}, ``Accelerated evaluation of automated
  vehicles in car-following maneuvers,'' \emph{IEEE Transactions on Intelligent
  Transportation Systems}, 2017.

\bibitem{halton1960efficiency}
J.~H. Halton, ``On the efficiency of certain quasi-random sequences of points
  in evaluating multi-dimensional integrals,'' \emph{Numerische Mathematik},
  vol.~2, no.~1, pp. 84--90, 1960.

\bibitem{tuncali2016utilizing}
C.~E. Tuncali, T.~P. Pavlic, and G.~Fainekos, ``Utilizing s-taliro as an
  automatic test generation framework for autonomous vehicles,'' in \emph{2016
  IEEE 19th International Conference on Intelligent Transportation Systems
  (ITSC)}.\hskip 1em plus 0.5em minus 0.4em\relax IEEE, 2016, pp. 1470--1475.

\bibitem{tuncali2018simulation}
C.~E. Tuncali, G.~Fainekos, H.~Ito \emph{et~al.}, ``Simulation-based
  adversarial test generation for autonomous vehicles with machine learning
  components,'' in \emph{2018 IEEE Intelligent Vehicles Symposium (IV)}.\hskip
  1em plus 0.5em minus 0.4em\relax IEEE, 2018, pp. 1555--1562.

\bibitem{prakash2021multi}
A.~Prakash, K.~Chitta, and A.~Geiger, ``Multi-modal fusion transformer for
  end-to-end autonomous driving,'' in \emph{Proceedings of the IEEE/CVF
  Conference on Computer Vision and Pattern Recognition}, 2021.

\bibitem{chen2021worldonrails}
D.~Chen, V.~Koltun, and P.~Kr{\"a}henb{\"u}hl, ``Learning to drive from a world
  on rails,'' in \emph{Proceedings of the IEEE/CVF International Conference on
  Computer Vision}, 2021, pp. 15\,590--15\,599.

\bibitem{dejanovic2017textx}
I.~Dejanovi{\'c}, R.~Vaderna, G.~Milosavljevi{\'c} \emph{et~al.}, ``Textx: a
  python tool for domain-specific languages implementation,''
  \emph{Knowledge-Based Systems}, vol. 115, pp. 1--4, 2017.

\bibitem{adapter}
O.~Hummel and C.~Atkinson, ``The managed adapter pattern: Facilitating glue
  code generation for component reuse,'' in \emph{Formal Foundations of Reuse
  and Domain Engineering}, S.~H. Edwards and G.~Kulczycki, Eds.\hskip 1em plus
  0.5em minus 0.4em\relax Springer Berlin Heidelberg, 2009, pp. 211--224.

\bibitem{carla-challenge}
\BIBentryALTinterwordspacing
O.~CARLA, ``Carla leaderboard challenge,'' 2020. [Online]. Available:
  \url{https://leaderboard.carla.org/}
\BIBentrySTDinterwordspacing

\bibitem{hartsell2019cps}
\BIBentryALTinterwordspacing
C.~Hartsell, N.~Mahadevan, S.~Ramakrishna \emph{et~al.}, ``Model-based design
  for cps with learning-enabled components,'' in \emph{Proceedings of the
  Workshop on Design Automation for CPS and IoT}, ser. DESTION '19.\hskip 1em
  plus 0.5em minus 0.4em\relax New York, NY, USA: Association for Computing
  Machinery, 2019, p. 1–9. [Online]. Available:
  \url{https://doi.org/10.1145/3313151.3313166}
\BIBentrySTDinterwordspacing

\end{thebibliography}
